\documentclass[runningheads]{llncs}

\usepackage{eccv}

\usepackage{eccvabbrv}

\usepackage{graphicx}
\usepackage{booktabs}
\usepackage{float}

\usepackage[accsupp]{axessibility}

\usepackage{hyperref}
\usepackage{multirow}
\usepackage{marvosym}
\usepackage{orcidlink}

\usepackage{xcolor}
\begin{document}
\raggedbottom
\setlength{\textfloatsep}{6pt plus 1pt minus 2pt}
\setlength{\floatsep}{6pt plus 1pt minus 2pt}
\setlength{\intextsep}{6pt plus 1pt minus 2pt}

\title{MedRepBench: Benchmarking Structured Understanding of Medical Report Images}

\titlerunning{MedRepBench}

\author{Fangxin Shang\inst{1} \and
Yuan Xia\inst{1}\textsuperscript{\Letter} \and
Dalu Yang\inst{1} \and
Yahui Wang\inst{1} \and
Binglin Yang\inst{1}}

\authorrunning{F.~Shang et al.}

\institute{Baidu Inc., China}

\maketitle
\begingroup
\renewcommand{\thefootnote}{\Letter}
\footnotetext{Corresponding author. Email: \email{babyxiayuan@gmail.com}}
\endgroup

\begin{abstract}
Medical report understanding from real-world document images is essential for generating patient-facing explanations and enabling structured information exchange in clinical systems.
Existing VLMs and LLMs have shown strong performance on document understanding, but structured understanding of medical reports remains insufficiently benchmarked. Therefore, we introduce MedRepBench, a benchmark with 1,925 de-identified Chinese medical report images spanning diverse departments, patient demographics, and acquisition formats.
In MedRepBench, we mainly focus on report-grounded interpretation rather than evaluating diagnostic reasoning, treatment recommendation, or the integration of patient history. The \textit{interpretation} is defined as structured extraction of report fields (e.g., \textit{item}, \textit{value}, \textit{unit}, \textit{reference range}, \textit{abnormal flag}) plus a patient-facing explanation grounded strictly in the report content.
The benchmark primarily evaluates end-to-end VLMs, and also includes a controlled text-only setting (high-quality OCR + LLM) to approximate an upper bound when character recognition errors are minimized.
Our evaluation framework provides two complementary protocols: (1) an objective protocol measuring field-level recall of structured items, and (2) an automated subjective protocol that uses an LLM-based judge to score factuality, interpretability, and reasoning quality under a fixed prompt.
Using the objective metric as a reward signal, we also provide a lightweight GRPO-based alignment baseline for a mid-sized VLM, which improves field-level recall by up to 6\%.
Finally, we analyze practical limitations of OCR+LLM pipelines, including layout-related errors and additional system latency, showing the need for robust end-to-end vision-based medical report understanding.
The dataset and evaluation resources are publicly available on \href{https://huggingface.co/datasets/MedRepBench/MedRepBench}{HuggingFace}.
\end{abstract}

\section{Introduction}
\label{sec:intro}

Medical report interpretation is an essential capability in modern healthcare, serving both patients and healthcare professionals. It is also a key component of AI-powered medical consultation system~\cite{Xia2020GenerativeAR,xia-etal-2022-speaker,Chen2024CoDTA}.
For patients, it transforms complex clinical documents into user-friendly explanations that support informed decision-making and promote healthier lifestyles outside clinical settings.
For physicians and healthcare institutions, it enables efficient electronic information exchange across disparate systems, improving referral workflows and multi-institutional collaboration.
As healthcare systems increasingly adopt electronic documentation, the demand for automated and scalable medical report interpretation solutions continues to grow.

\begin{figure}[htbp]
  \centering
  \includegraphics[width=0.7\columnwidth]{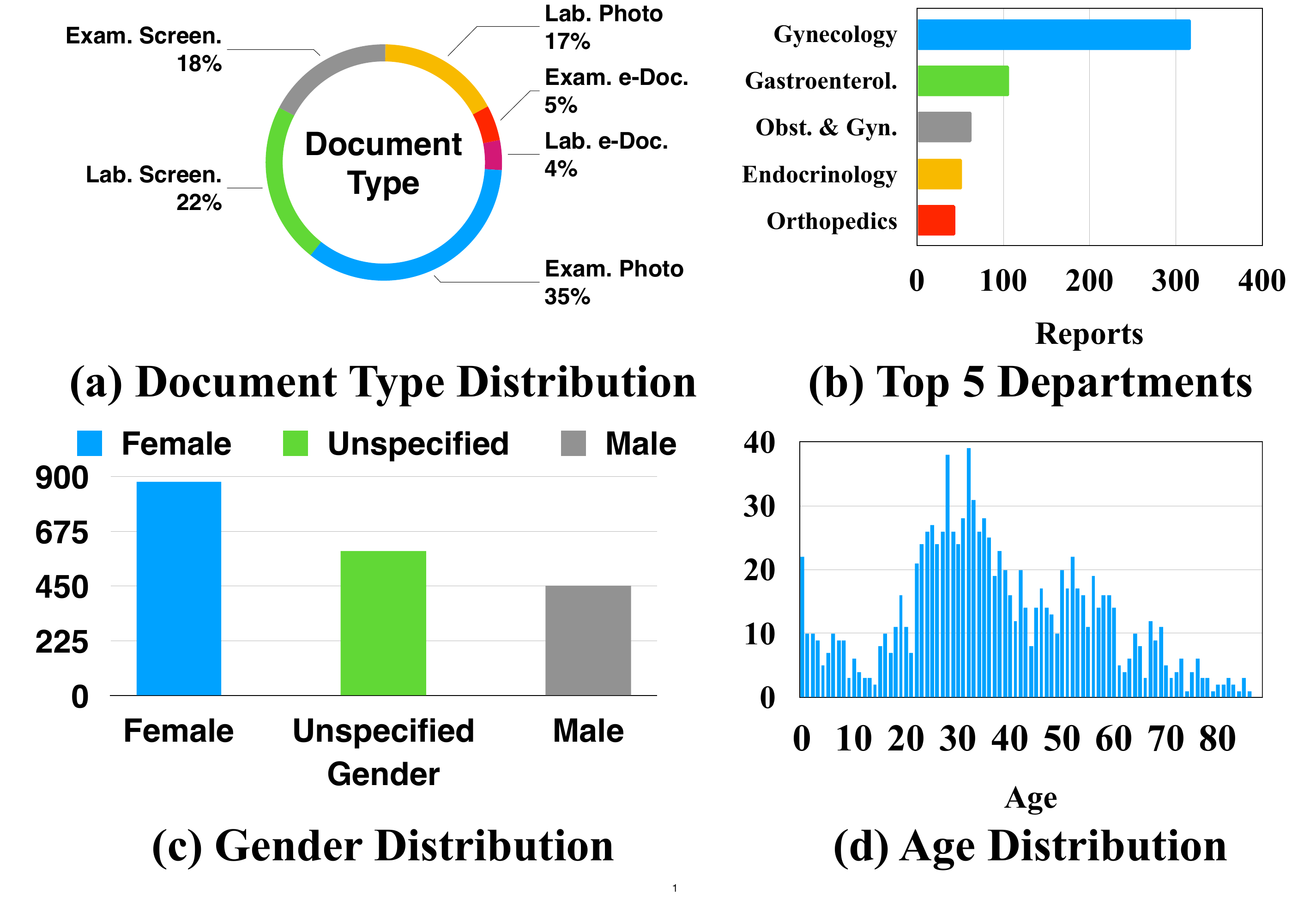}
  \caption{
  Overview of the MedRepBench dataset.
(a) Joint distribution of document types (Exam./Lab.) and acquisition methods (Photo, Screenshot, e-Doc.);
(b) Report counts for the top-5 clinical departments;
(c) Patient gender distribution (with \textit{Unspecified} denoting missing metadata);
(d) Patient age distribution across the corpus.}
  \label{fig:medrepbench}
\end{figure}

Unlike medical report generation~\cite{hartsock2024vision}, which focuses on producing narrative-style descriptions, medical report interpretation aims to extract structured clinical findings~\cite{moon2025lunguage} and generate patient-facing explanations grounded in factual content.
This task sits at the intersection of information extraction and layperson-targeted summarization, supporting applications such as at-home health monitoring, physician-patient communication, and intelligent referral systems.

Although both vision--language models (VLMs) and large language models (LLMs) demonstrate strong capabilities in document understanding, prior work~\cite{liu2024medbench} has not adequately evaluated their performance on structured interpretation of medical reports.
We therefore focus on assessing the end-to-end capability of VLMs to extract and explain structured clinical content directly from raw report images, and systematically compare them with OCR-assisted pipelines (OCR+VLM and OCR+LLM).
From an application standpoint, OCR-based pipelines remain the most accurate and cost-effective option whenever high-quality text can be reliably extracted from medical reports.
However, they introduce an additional subsystem that must be engineered and maintained for each domain and document template, and they strip away layout and visual cues from the downstream LLM, leading to architectural overhead and error propagation.

In parallel, emerging vision language models with text-token-free visual encoders ~\cite{wei2025deepseek, cheng2025glyphscalingcontextwindows}, which encode written content directly as visual tokens, challenge the traditional \textit{OCR-then-LLM} decomposition.
These models call for benchmarks that stress end-to-end visual understanding and structured prediction directly from images, beyond purely text-based extraction.
MedRepBench is therefore designed to compare OCR-assisted and no-OCR paradigms under a unified protocol and to serve as a concrete reference point and diagnostic signal for future text-token-free VLMs.
Our benchmark reveals that VLMs, although still lagging behind strong text-only models under OCR-assisted settings, remain a promising direction for unified, layout-aware medical document interpretation.

The medical report interpretation task presents unique complexities compared to general document understanding: (1) \textbf{diverse data acquisition methods}, such as handheld device photography, PDF documents, and mobile web screenshots; (2) \textbf{high variability in image quality}, including occlusions, folds, partial captures, and illumination issues; and (3) \textbf{heterogeneous layout styles}, as medical reports lack universally enforced formatting standards.
These factors collectively demand robust model performance across diverse and noisy input scenarios.

To address these challenges, we introduce \textbf{MedRepBench}, a benchmark designed to evaluate modern models in medical report interpretation.
MedRepBench consists of 1,925 de-identified real-world medical reports spanning multiple clinical departments and patient demographics.
Each report includes metadata such as document type (e.g., \emph{Exam. Report}, \emph{Lab Report}), department (e.g., \emph{Gynecology}, \emph{Pediatrics}), patient gender, and age.

As shown in Figure~\ref{fig:medrepbench}, our proposed MedRepBench includes two primary document types---exam reports and lab reports---captured through diverse acquisition methods such as photos, screenshots, and electronic documents.
These reports originate from a wide range of medical institutions, covering substantial cross-institution layout variations, thus reflecting real-world heterogeneity in medical documentation.
Such heterogeneity, combined with broad departmental coverage and diverse patient demographics, is crucial for evaluating the robustness and generalizability of medical report interpretation models.

\textbf{Our main contributions are as follows:}
\begin{itemize}
    \setlength{\itemsep}{1pt}
    \setlength{\parsep}{0pt}
    \setlength{\topsep}{2pt}
    \item We introduce \textbf{MedRepBench}, a benchmark for \emph{vision-grounded structured understanding} of real-world Chinese medical report images, featuring 1,925 de-identified laboratory and examination reports.
    It supports a dual evaluation protocol: (1) \textit{objective} field-level recall over five attributes, and (2) \textit{automated subjective} evaluation of factuality, interpretability, and reasoning quality via an LLM judge.

    \item We benchmark representative open-source VLMs and LLMs under both end-to-end (image-only) and OCR-assisted settings, revealing current strengths, failure modes, and the remaining gap in the no-OCR configuration.

    \item We provide a lightweight GRPO-based alignment baseline that uses field-level recall as a reward, showing up to \textbf{+6\%} absolute recall improvement over SFT for a mid-scale VLM.
\end{itemize}

The dataset, benchmark protocol, and evaluation resources are released on \href{https://huggingface.co/datasets/MedRepBench/MedRepBench}{HuggingFace}, including de-identified report images, structured metadata/labels, documentation, prompt templates, and evaluation scripts for reproducible evaluation.
The dataset is released under the CC BY-NC 4.0 license.

\begin{figure}[H]
  \centering
  \includegraphics[width=0.86\textwidth]{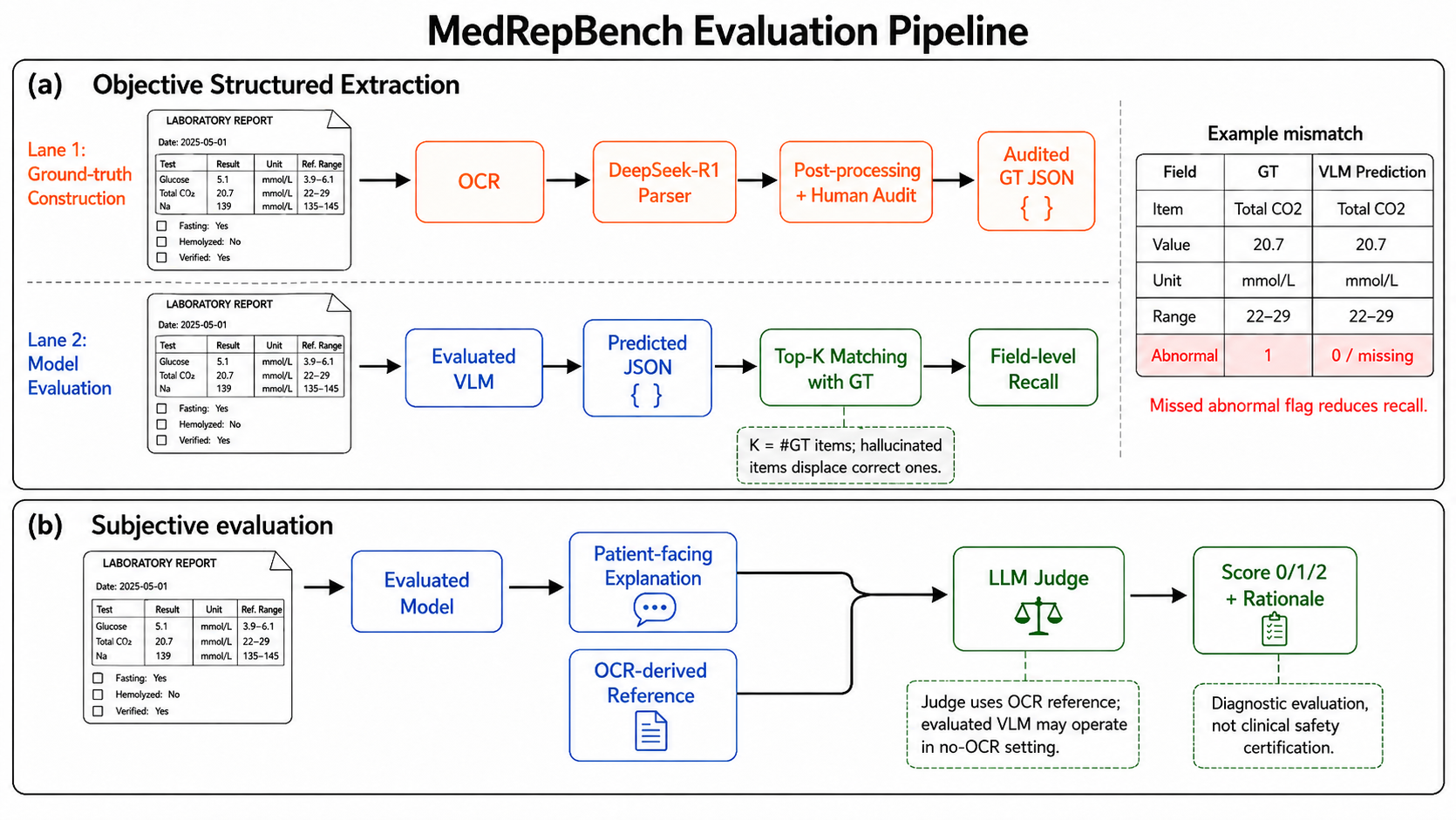}
  \caption{
Evaluation pipeline and qualitative comparison in MedRepBench.
(a) Objective evaluation separates OCR-assisted ground-truth construction from VLM prediction and Top-K field-level recall; the inset shows a VLM--GT abnormal-flag mismatch.
(b) Subjective evaluation scores patient-facing explanations with an LLM judge using OCR-derived reference content.
  }
  \label{fig:task_examples}
\end{figure}

\section{The Proposed Benchmark}
\subsection{Dataset Construction}
Figure~\ref{fig:task_examples} provides an overview of the evaluation pipeline, covering objective structured extraction and subjective scoring of patient-facing explanations. We collect medical report images from a private, real-world online medical service source spanning multiple clinical departments and document types, including handheld photographs, PDFs, and mobile screenshots.
The reports were uploaded in practical service scenarios, and their secondary use for benchmark construction is governed by the service's user agreement and institutional compliance requirements.
We randomly sample 8,000 reports and apply a multi-stage curation pipeline covering de-identification, quality filtering, deduplication, and content validation, resulting in 1,925 high-quality images.

To obtain structured labels, we first extract text with a proprietary OCR system optimized for medical documents.
The OCR text is then parsed into structured JSON by DeepSeek-R1~\cite{guo2025deepseek} under a constrained schema, producing item-level fields such as \textit{name}, \textit{value}, \textit{unit}, \textit{reference range}, and \textit{abnormality flag}, as well as document-level metadata including \textit{document type}, \textit{department}, \textit{gender}, and \textit{age}.

\textbf{Filtering and Quality Control.}
We remove low-quality or incomplete captures using heuristic quality filters (e.g., short-side resolution below 800 px, severe motion blur, and truncated pages).
We then manually inspect a random subset of 500 samples to verify the filtering results.
We then deduplicate reports by combining perceptual image hashing with OCR text similarity, discarding 12.3\% near-duplicates.
Next, we perform content validation to ensure each retained report contains at least one structured laboratory panel with valid reference ranges.
To control bootstrapping noise, we enforce schema validation and rule-based normalization (e.g., unit standardization, numeric formatting, and reference-range pattern checks), and further audit a random subset of 200 reports with human annotators.
We observe over 97\% field-level agreement; the main residual errors involve nested-table field misalignment and OCR errors on handwritten annotations, and all errors found during the audit are corrected before release.

\textbf{De-identification and Release.}
Before release, we apply OCR-based personally identifiable information (PII) detection for names, addresses, ID numbers, and other direct identifiers, followed by region-of-interest pixel blackout.
All 1,925 images are then manually inspected to ensure that no individual can be traced from the released data.
The public \href{https://huggingface.co/datasets/MedRepBench/MedRepBench}{HuggingFace repository} contains the de-identified images under a censored image directory and a metadata/label CSV file with document-level attributes and structured item annotations.

Finally, we emphasize that our main benchmark conclusions rely on \emph{field-level recall} against the released structured labels, rather than any LLM-judge component used elsewhere in the evaluation framework.

\subsection{Structured Interpretation Evaluation}

Previous benchmarks for report evaluation---such as BLEU, ROUGE, or CIDEr---focus on surface-level text similarity, which is insufficient for assessing structured information extraction.
MedRepBench instead introduces a field-level evaluation protocol based on recall, defined for each field \( f \) as:
\begin{equation}
\mathrm{Recall}_f = \frac{\text{\# correctly extracted field } f}{\text{\# ground-truth field } f}.
\label{eq:field_recall}
\end{equation}

We evaluate recall over five core fields as mentioned above.
The primary objective metric is the average of field-wise recall across all matched items:
\begin{equation}
\mathrm{AverageRecall} = \frac{1}{|\mathcal{F}|} \sum_{f \in \mathcal{F}} \mathrm{Recall}_f,
\label{eq:avg_recall}
\end{equation}
where \( \mathcal{F} \) denotes the set of target fields and \( |\mathcal{F}| = 5 \) in MedRepBench.

To avoid rewarding unconstrained over-generation, we truncate each model output to Top-K items, where \(K\) is the number of ground-truth items in the corresponding report.
Under this protocol, hallucinated or irrelevant items compete for the same Top-K slots and can displace correct items, directly reducing field-level recall.
Thus, although the primary metric is recall, the fixed-budget matching protocol implicitly penalizes spurious generations.

\subsection{Automated Subjective evaluation via LLM}
In addition to field-level recall, we use an LLM-based evaluator to assess the interpretability and factuality of model outputs.
The evaluator is prompted with the model-generated explanation and OCR-derived reference content, returning a scalar score (e.g., 0--2) and a brief rationale. This subjective score complements the objective metric by evaluating whether the output forms a report-grounded and internally consistent patient-facing explanation.

In our setting, \textit{factuality} measures whether the generated explanation is strictly supported by evidence in the report, i.e., it is consistent with the extracted structured fields and does not introduce non-existent test items, values, units, or reference ranges.
\textit{Interpretability} reflects whether the explanation is understandable to lay users by organizing results clearly (e.g., highlighting key abnormal items), using accessible language, and avoiding unsupported diagnoses or treatment recommendations.
\textit{Reasoning quality} assesses whether abnormality and key takeaways are derived from value--range comparisons or discrete outcomes (e.g., positive/negative), without internal contradictions.

While DeepSeek-R1 is used for both initial annotation and as the LLM judge, we decouple these roles. Model performance conclusions are based on objective field-level recall, independent of DeepSeek-R1's judgment. DeepSeek-R1 annotations are post-processed and verified by human reviewers to minimize systematic errors. The LLM-as-judge scores are primarily for ranking and qualitative comparison, further validated through a pilot human study in Section~4.2.

\subsection{Reinforcement Learning Optimization}

To improve structured item extraction, we adopt \textbf{Group Relative Policy Optimization (GRPO)}~\cite{shao2024deepseekmath,lai2025medr1,pan2025medvlm} to align a vision--language model with our benchmark objective, using field-level recall as a non-differentiable reward signal.
This component is included as a practical instantiation of recent RL-based alignment techniques in the context of medical report understanding, and we do \emph{not} claim algorithmic novelty.

\paragraph{Reward Design and Optimization.}
We treat report interpretation as sequence generation: given an image or OCR-assisted prompt, the model outputs a JSON list of five-field items (\textit{name}, \textit{value}, \textit{unit}, \textit{range}, \textit{abnormality flag}).
The GRPO reward is the average field-level recall over target fields \( \mathcal{F} \), where \( |\mathcal{F}| = 5 \):
\begin{equation}
R = \frac{1}{|\mathcal{F}|} \sum_{f \in \mathcal{F}} \mathrm{Recall}_f.
\label{eq:reward}
\end{equation}
Recall is computed between one-to-one matched prediction--reference pairs aligned by normalized item names.
Following GRPO, candidate outputs in each group are scored by \(R\), transformed into group-relative advantages, and optimized with a PPO-style clipped objective with KL regularization to the SFT reference policy.

\paragraph{Implementation Details.}
We apply GRPO on top of an InternVL3-8B model~\cite{zhu2025internvl3} that has been previously fine-tuned via supervised learning (SFT) on a separate dataset for the subjective interpretation task.
During GRPO training, the vision encoder and image-to-text projector are frozen to preserve stable multimodal alignment.
We insert LoRA~\cite{hu2022lora} modules (rank=128) into all attention layers of the LLM (query, key, value projections), and only LoRA weights are updated.
Training is performed for 2 epochs on a curated dataset of 800 structured report samples following the MedRepBench format.

This lightweight optimization yields a +6\% gain in average field-level recall over the SFT baseline, along with improved subjective interpretability (see Section~\ref{subsec:grpo_baseline} for details).

\section{Evaluation}

\subsection{Models and Experimental Setup}

To ensure reproducibility and privacy compliance in clinical applications, we evaluate exclusively on publicly available vision-language models (VLMs) and large language models (LLMs). Proprietary frontier APIs (e.g., GPT, Gemini, and Claude) are excluded.

\textbf{VLMs}: We include a broad selection of recent vision-language models capable of document-level understanding, including InternVL2.5~\cite{chen2024expanding}, InternVL3~\cite{zhu2025internvl3}, Qwen2.5-VL~\cite{bai2025qwen2}, and LLaMA-4~\cite{meta2025llama}.
These models are evaluated using their image input interface under consistent inference settings.

We intentionally focus on general-domain VLMs to evaluate their transferability from generic document and vision-language understanding tasks to structured medical report extraction. Extending the benchmark to medical-specialized VLMs is a natural direction for future leaderboard development.

\textbf{LLMs}: We additionally benchmark several text-only LLMs, including Qwen3 series~\cite{yang2025qwen3} and the DeepSeek-R1-Distill~\cite{guo2025deepseek}. For these models, OCR-extracted text is provided as part of the input prompt. DeepSeek-R1 is also employed as an automatic judge for subjective evaluation.

For objective evaluation, we consider two settings: (1) end-to-end VLM interpretation from raw images, and (2) an auxiliary OCR-assisted setup where recognized text is provided to both VLMs and LLMs.
In subjective evaluation, only the evaluator (DeepSeek-R1) receives OCR input, while models generate interpretations from their native inputs (images for VLMs, OCR for LLMs).
This design ensures interpretability assessment is decoupled from access to ground-truth text, while remaining anchored to a consistent reference.

Although MedRepBench supports both visual and text-based pipelines, its primary goal is to evaluate the end-to-end capabilities of VLMs.
OCR-assisted variants serve only to approximate upper bounds and diagnose non-OCR-related limitations.

\subsection{Objective and Subjective Evaluation Protocols}

We design two complementary evaluation protocols: objective evaluation, focusing on structured field extraction accuracy, and subjective evaluation, focusing on report-grounded factuality, readability, and reasoning quality.

\paragraph{Objective Evaluation.}
This protocol is applied exclusively to laboratory-style reports (e.g., blood tests), where each report contains a list of structured test items represented as five-field tuples.
Each item contains the test name, measured value, reference range, measurement unit, and an abnormality flag.
Models are required to output these as a JSON array, as illustrated in Figure~\ref{fig:task_examples}.
We do not apply this objective protocol to examination reports because they are predominantly narrative and often lack a stable itemized schema; forcing field-level labels for such reports would introduce unreliable pseudo-labels.
Instead, examination reports are covered by the subjective protocol, which evaluates report-grounded factuality, reasoning, and patient-facing clarity.

For each item, we define five target fields: name, value, unit, reference\_range, and is\_abnormal. The first four are directly extractable from the report, while is\_abnormal is a derived label that requires comparing the measured value with the reference range and handling discrete labels such as 'positive/negative' or semi-quantitative levels.
We expect models to infer is\_abnormal rather than copying it from the text, and we treat it as a reasoning-oriented sub-task.

Performance is evaluated by comparing the model-generated JSON with the ground-truth JSON and measuring field-level recall.
An item is matched to a reference only if the \textit{name} field matches exactly (after normalization).
Once matched, each ground-truth item is considered once. Recall is then computed for the remaining fields independently: \textit{value}, \textit{unit}, \textit{range}, and \textit{abnormal}.
To account for minor formatting variation (e.g., 2--4 vs. 2$\sim$4), fuzzy matching is permitted for the \textit{range} field, while other fields require exact equivalence (either numerical or string).

Preprocessing includes whitespace trimming, irrelevant token removal, and fault-tolerant JSON parsing to ensure consistent evaluation.
We report both per-field recall and averaged recall across all five attributes.

\paragraph{Subjective Evaluation.}
In this setting, models are asked to generate human-readable, report-grounded explanations based on the input report.
For consistency, all models are evaluated under the no-OCR configuration, where the recognized text is not provided to the model.

We employ DeepSeek-R1 as the evaluator model, prompted to assess three key dimensions: factual accuracy, reasoning validity, and patient-facing clarity.
Each interpretation is scored on a discrete 3-point scale:

\begin{itemize}
\item \textbf{0}: Severe factual errors or unsupported clinical claims, such as inventing absent findings, reversing abnormality status, or giving diagnosis/treatment advice not supported by the report.
\item \textbf{1}: Minor inaccuracies or stylistic issues, without major unsupported clinical advice.
\item \textbf{2}: Accurate, well-structured, and clearly grounded in the report.
\end{itemize}

We compute two metrics: \textbf{acceptability rate} (proportion of samples with score $\geq 1$), and \textbf{excellence rate} (proportion with score 2).
To ensure robustness, all evaluations use deterministic decoding (temperature=0). A subset of responses is re-evaluated across multiple seeds to confirm evaluation stability.

\section{Results}

\subsection{Objective Evaluation (Structured Field Recall)}

We evaluate structured extraction on laboratory-style reports using field-level recall over five attributes: \textit{name}, \textit{value}, \textit{unit}, \textit{reference range}, and \textit{abnormality}.
Predicted entries are matched to ground truth via normalized \textit{name} alignment, after which recall is computed independently for the remaining attributes.
Results are averaged over three decoding runs with different random seeds; standard deviations are below 5\%, indicating stable evaluation.

Table~\ref{tab:recall_vlm_with_vs_without_ocr} reports VLM performance in our primary \textit{no-OCR} setting (end-to-end, vision-grounded extraction) and the corresponding \textit{OCR-assisted} setting for reference.
Several models, notably the InternVL3 family and Qwen2.5-VL-32B, achieve meaningful recall even without explicit text input.
Providing OCR text consistently improves recall across models and fields---especially for \textit{value} and \textit{reference range}---and serves as an approximate upper bound when visual perception errors are minimized.
Nevertheless, as discussed later, OCR-driven pipelines remain vulnerable to cascading errors and are less grounded in layout, motivating further progress toward robust end-to-end VLMs.

\begin{table}[!t]
\caption{Structured field-level recall (\%) of VLMs on laboratory-style reports. no-OCR denotes end-to-end image-only inference; OCR-asst. denotes OCR-assisted inference where recognized text is provided as additional input.}
\centering
\resizebox{\linewidth}{!}{
\begin{tabular}{l cc cc cc cc cc}
\toprule
\multirow{2}{*}{Model} &
\multicolumn{2}{c}{$R_{name}\uparrow$} &
\multicolumn{2}{c}{$R_{value}\uparrow$} &
\multicolumn{2}{c}{$R_{unit}\uparrow$} &
\multicolumn{2}{c}{$R_{range}\uparrow$} &
\multicolumn{2}{c}{$R_{abnormal}\uparrow$} \\
\cmidrule(lr){2-3}\cmidrule(lr){4-5}\cmidrule(lr){6-7}\cmidrule(lr){8-9}\cmidrule(lr){10-11}
& no-OCR & OCR-asst.
& no-OCR & OCR-asst.
& no-OCR & OCR-asst.
& no-OCR & OCR-asst.
& no-OCR & OCR-asst. \\
\midrule
InternVL2.5-8B      & 71.38 & 85.64 & 55.71 & 70.14 & 60.63 & 77.85 & 58.88 & 79.40 & 45.06 & 58.59 \\
InternVL2.5-38B     & 84.83 & 95.20 & 68.85 & 85.02 & 74.93 & 89.00 & 72.37 & 89.44 & 66.01 & 81.52 \\
InternVL3-8B        & 88.21 & 95.25 & 71.29 & 84.87 & 75.56 & 88.00 & 70.63 & 88.80 & 67.92 & 77.97 \\
InternVL3-14B       & 88.70 & 95.25 & 67.69 & 86.53 & 74.63 & 89.52 & 67.96 & 88.81 & 70.12 & 83.56 \\
InternVL3-38B       & 88.72 & 94.69 & 64.83 & 85.64 & 75.63 & 88.79 & 69.36 & 89.88 & 66.74 & 77.55 \\
Qwen2.5-VL-7B       & 89.90 & 93.94 & 73.17 & 82.46 & 77.31 & 81.61 & 78.33 & 85.07 & 68.36 & 76.39 \\
Qwen2.5-VL-32B      & 90.13 & 95.89 & 76.32 & 84.72 & 79.30 & 87.71 & 81.61 & 89.15 & 59.66 & 83.95 \\
LLaMA-4-Scout       & 87.90 & 92.68 & 65.69 & 78.86 & 74.29 & 82.94 & 67.87 & 77.91 & 63.24 & 69.23 \\
LLaMA-4-Maverick    & 88.51 & 96.68 & 72.26 & 87.35 & 78.83 & 90.23 & 77.58 & 90.74 & 75.83 & 86.56 \\
\bottomrule
\end{tabular}
}

\label{tab:recall_vlm_with_vs_without_ocr}
\end{table}

Although the OCR+LLM pipeline performs well on both objective and subjective metrics, its performance remains highly dependent on OCR quality.
Recognition errors, such as missing entries, structural misalignment, and incorrect segmentation, can propagate through the pipeline and affect downstream reasoning, resulting in factual hallucinations such as incorrectly identifying normal findings as abnormal.
Such cascading errors are hard to correct due to the non-end-to-end nature of the pipeline.
Furthermore, by discarding visual layout cues, OCR+LLM models lack spatial grounding, limiting their robustness and clinical reliability in real-world settings.

\begin{table}[!t]
\caption{Field-level recall (\%) for LLMs under OCR-assisted input.}
\label{tab:recall_with_ocr_llm}
\centering
\begin{tabular*}{\linewidth}{@{\extracolsep{\fill}}lccccc}
\toprule
Model & $R_{name}\uparrow$ & $R_{value}\uparrow$ & $R_{unit}\uparrow$ & $R_{range}\uparrow$ & $R_{abnormal}\uparrow$ \\
\midrule
Qwen3-1.7B & 92.88 & 70.50 & 79.42 & 80.91 & 69.04 \\
Qwen3-4B & 94.64 & 74.59 & 86.68 & 87.40 & 64.29 \\
Qwen3-8B & 94.93 & 79.48 & 88.36 & 88.26 & 68.50 \\
Qwen3-14B & 95.75 & 87.43 & 89.69 & 90.18 & 85.43 \\
Qwen3-32B & 95.66 & 87.46 & 90.92 & 90.68 & 85.77 \\
Qwen3-30B-A3B & 96.08 & 86.48 & 88.88 & 90.08 & 82.67 \\
Qwen3-235B-A22B & 94.96 & 87.27 & 89.09 & 90.56 & 84.54 \\
Qwen3-235B-A22B-2507 & 96.42 & 89.40 & 91.14 & 92.48 & 88.03 \\
DeepSeek-R1-Qwen-7B & 91.68 & 70.60 & 75.83 & 75.13 & 67.13 \\
DeepSeek-R1-Qwen-14B & 95.50 & 84.61 & 88.07 & 88.38 & 83.01 \\
DeepSeek-R1-Qwen-32B & 92.93 & 83.40 & 86.36 & 87.53 & 82.29 \\
DeepSeek-V3 & 96.48 & 88.83 & 90.75 & 92.60 & 89.06 \\
\bottomrule
\end{tabular*}
\end{table}

\begin{table}[!t]
\caption{LLM-based subjective evaluation scores under no-OCR input.}
\label{tab:subjective_without_ocr}
\centering
\begin{tabular*}{\linewidth}{@{\extracolsep{\fill}}lcc}
\toprule
Model & Accept. $\uparrow$ & Excel. $\uparrow$ \\
\midrule
InternVL2.5-8B & 33.67 & 17.30 \\
InternVL2.5-38B & 57.97 & 42.45 \\
InternVL3-8B & 48.24 & 31.83 \\
InternVL3-14B & 47.00 & 28.55 \\
InternVL3-38B & 51.15 & 32.70 \\
Qwen2.5-VL-7B & 26.21 & 5.58 \\
Qwen2.5-VL-32B & 67.83 & 51.40 \\
LLaMA-4-Scout & 46.84 & 26.18 \\
LLaMA-4-Maverick & 60.78 & 42.74 \\
\bottomrule
\end{tabular*}
\end{table}

\begin{figure}[htbp]
    \centering
    \includegraphics[width=1.0\columnwidth]{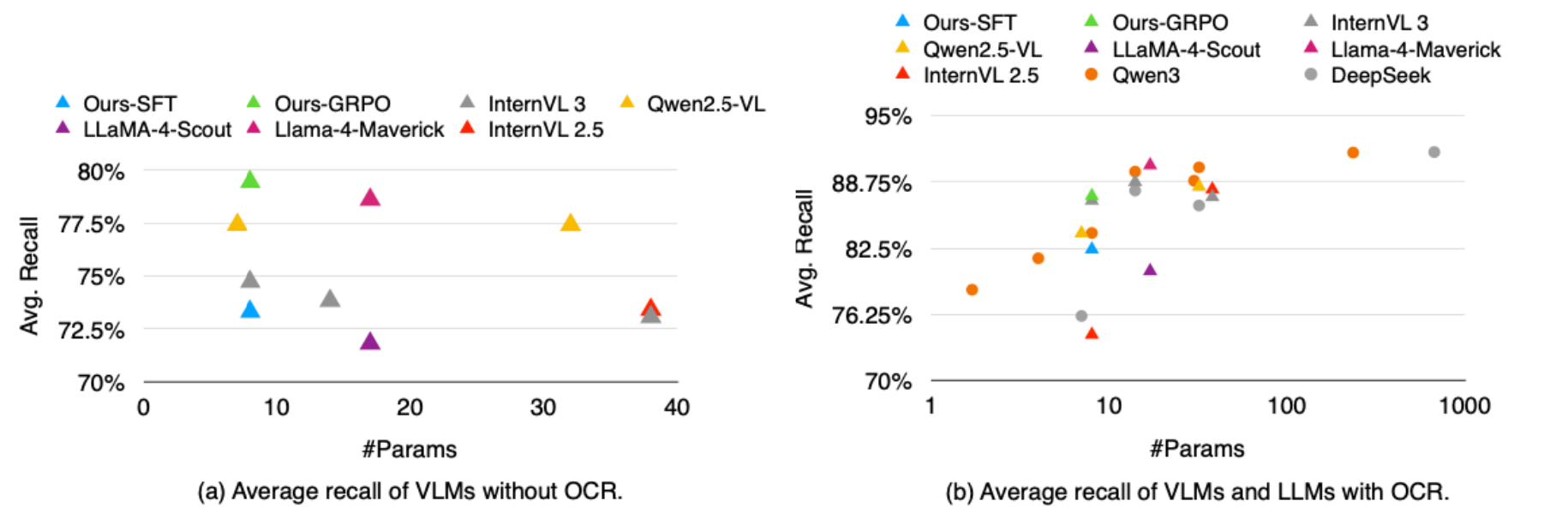}
    \caption{Average field-level recall (\%) vs. model size.
(a) no-OCR input; (b) OCR-assisted input.
VLMs (\(\triangle\)) and LLMs (\(\circ\)) are distinguished by markers.
Ours-GRPO substantially improves 8B VLM performance, and is competitive with larger baselines in the no-OCR setting. Y-axis in (a) is limited to 70--80\% for clarity; InternVL2.5-8B (58.33\%) is excluded.}
    \label{fig:recall_curve}
\end{figure}

Figure~\ref{fig:recall_curve} visualizes average recall across all five fields with respect to model size.
In both no-OCR and OCR-assisted settings, performance generally scales with model capacity.
However, several models deviate from this trend, suggesting that alignment strategies and training data play crucial roles.
Notably, our RL-optimized model (Ours-GRPO) outperforms much larger baselines under no-OCR conditions, highlighting the potential of benchmark-driven optimization.

These results confirm the difficulty of end-to-end report interpretation, especially in the absence of OCR.
On average, field-level recall drops by 10--20\% when OCR input is removed.
This gap reveals substantial headroom for improving visual-text alignment and structured reasoning in VLMs, further motivating the need for targeted benchmarks such as MedRepBench.

\paragraph{No-OCR Failure Modes.}
Manual inspection shows three recurring error patterns in end-to-end VLM outputs: (1) character-level mistakes on fine-print numeric values and units, (2) layout misalignment in nested tables or multi-column reports, and (3) missed abnormality flags when visual cues are subtle or separated from the corresponding value.
These failures explain why OCR-assisted pipelines remain strong while also motivating models that can jointly preserve text, layout, and visual grounding.

\subsection{Subjective Evaluation (Interpretability Score)}

We additionally conduct a pilot human study to validate the reliability of the LLM-based subjective scores.
Beyond factual field extraction, we evaluate models from the perspective of clinical usability: each model-generated explanation is scored by DeepSeek-R1 along three dimensions---factuality, reasoning quality, and patient-facing clarity.
The scores are mapped to a three-level scale, and we report two summary metrics: \textbf{Acceptability Rate} (scores $\geq 1$) and \textbf{Excellence Rate} (scores = 2).

Table~\ref{tab:subjective_without_ocr} presents the subjective evaluation results in the no-OCR setting. Larger VLMs such as Qwen2.5-VL-32B and LLaMA-4-Maverick generally achieve stronger acceptability and excellence rates, while smaller or insufficiently aligned models often generate vague or factually inaccurate interpretations.
These results highlight the need for visual grounding and domain-aware alignment in patient-facing medical report explanations.

\paragraph{Agreement Between LLM-Based Evaluator and Human Experts.}

To assess the reliability of our LLM-based evaluator, we conduct a consistency study against expert human judgments. We randomly sample 20 cases from each of the three rating categories (score 0, 1, and 2) from the 1,925 evaluation set, resulting in a total of 60 samples. Each case is independently rated by three experienced annotators (licensed medical reviewers or clinically trained annotators). Final human scores are determined via majority voting. If consensus could not be reached, the case was discarded and replaced with another randomly sampled instance from the same score category.

\begin{figure}[tbp]
  \centering
  \includegraphics[width=0.5\columnwidth]{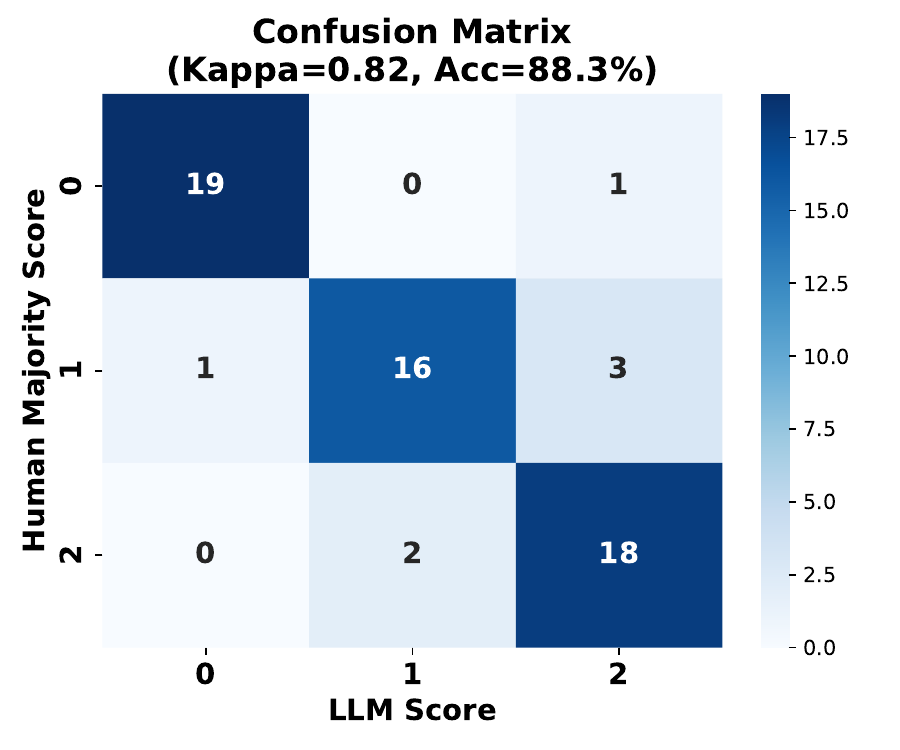}
  \caption{Confusion matrix between LLM-based evaluator scores and human expert majority vote. High agreement is observed (Cohen's $\kappa$ = 0.82, Accuracy = 88.3\%).}
  \label{fig:confusion_llm_human}
\end{figure}

Figure~\ref{fig:confusion_llm_human} shows the confusion matrix between the LLM-based evaluator and the aggregated human scores. Overall agreement is high, with an accuracy of \textbf{88.3\%} and Cohen's $\kappa$ of \textbf{0.82}, suggesting substantial consistency under our evaluation rubric. Most disagreements occur between adjacent ratings (e.g., 1 vs.\ 2), indicating borderline cases rather than clear-cut factual contradictions.
We emphasize that this pilot study serves as a sanity check rather than a substitute for expert clinical assessment; the LLM judge is used as a \emph{diagnostic} and scalable signal for ranking and qualitative comparison, not as an independent clinical assessment.

\subsection{A Practical Alignment Baseline (GRPO)}
\label{subsec:grpo_baseline}

To reduce the performance gap in the \textit{no-OCR} setting, where vision-grounded structured extraction remains challenging even for strong VLMs, we further optimize a supervised fine-tuned InternVL3-8B model using GRPO.
In our GRPO experiments, the policy input follows the no-OCR setting (images only), and the reward is computed against the released JSON labels.
MedRepBench provides a rigorous objective protocol and an explicit field-level recall signal, which naturally serves as a reward for this alignment.

We present GRPO here as a lightweight and reproducible baseline to demonstrate that MedRepBench induces a meaningful reward landscape for improving structured extraction; we do \emph{not} claim novelty of the underlying RL algorithm.

Concretely, we first perform 2 epochs of supervised fine-tuning (SFT) on InternVL3-8B using $\sim$15k human-annotated instruction--response pairs derived from real-world medical report interpretations, yielding \textbf{Ours-SFT}.
For reinforcement learning, we additionally curate 800 laboratory-report samples formatted according to our benchmark schema and run 2 epochs of GRPO with LoRA (rank 128), producing \textbf{Ours-GRPO}.

\begin{table}[htbp]
\caption{Field-level recall (\%) for Ours-GRPO and baselines.}
\label{tab:grpo_objective}
\centering
\begin{tabular*}{\linewidth}{@{\extracolsep{\fill}}lccc}
\toprule
Model & OCR & \#Param & Avg. Recall $\uparrow$ \\
\midrule
LLaMA-4-Maverick & No & 17B & 78.60 \\
Qwen2.5-VL-32B & No & 32B & 77.41 \\
Ours-SFT & No & 8B & 73.31 \\
Ours-GRPO & No & 8B & \textbf{79.45} \\ \hline
Qwen2.5-VL-7B & Yes & 7B & 83.89 \\
InternVL2.5-8B & Yes & 8B & 74.32 \\
Qwen3-8B & Yes & 8B & 83.90 \\
InternVL3-8B & Yes & 8B & 86.98 \\
Ours-SFT & Yes & 8B & 82.37 \\
Ours-GRPO & Yes & 8B & \textbf{87.41} \\
\bottomrule
\end{tabular*}
\end{table}

\begin{table}[htbp]
\caption{Subjective scores for Ours-GRPO and baselines.}
\label{tab:grpo_subjective}
\centering
\begin{tabular*}{\linewidth}{@{\extracolsep{\fill}}lccc}
\toprule
Model & Param & Accept. $\uparrow$ & Excel. $\uparrow$ \\
\midrule
Ours-SFT & 8B & 56.27 & 38.86 \\
Ours-GRPO & 8B & 60.64 & 42.45 \\
LLaMA-4-Maverick & 17B & 60.78 & 42.74 \\
Qwen2.5-VL-32B & 32B & \textbf{67.83} & \textbf{51.40} \\
InternVL2.5-38B & 38B & 57.97 & 42.45 \\
InternVL3-38B & 38B & 51.15 & 32.70 \\
\bottomrule
\end{tabular*}
\end{table}

Table~\ref{tab:grpo_objective} and Table~\ref{tab:grpo_subjective} compare \textbf{Ours-GRPO} against baselines in both objective and subjective settings, with Figure~\ref{fig:recall_curve} illustrating recall improvements under no-/with-OCR conditions.
In the no-OCR setting, \textbf{Ours-GRPO} achieves the highest average recall (79.45\%), outperforming LLaMA-4-Maverick (78.60\%) and Qwen2.5-VL-32B (77.41\%), and surpassing its supervised baseline (73.31\%) by +6\%.

Although \textbf{Ours-GRPO} is not the top-ranked model in subjective interpretability, its excellence rate of 42.45\% is on par with the much larger LLaMA-4-Maverick (42.74\%) and exceeds that of InternVL3-38B (32.70\%).
These results confirm that MedRepBench offers a meaningful reward landscape for policy optimization and that targeted reinforcement learning can further enhance structured interpretation, even beyond supervised baselines.

\section{Related Work}

\subsection{Medical Report Interpretation vs. Generation}
Most prior work in medical AI has focused on report generation or VQA using vision--language models (VLMs).
For example, Argus~\cite{liu2025argus} and MedVAG~\cite{arisoy2025medvag} tackle radiology report generation, while reviews~\cite{hartsock2024vision,arisoy2025medvag} highlight that structured field extraction---crucial for report \emph{interpretation}---is underexplored.
In contrast, MedRepBench specifically targets the interpretive task: extracting structured findings and evaluating readability rather than generating narrative summaries.

\subsection{Document Understanding and VQA Benchmarks}
Recent benchmarks in document understanding span both general and medical domains.
General-purpose benchmarks such as the MMDocBench~\cite{zhu2024mmdocbench} and BenchX~\cite{zhou2024benchx} focus on layout analysis, structure parsing, and multi-task performance across business or legal documents.
In the medical domain, visual question answering (VQA) benchmarks such as OmniMedVQA~\cite{hu2024omnimedvqa}, GEMeX~\cite{liu2024gemex}, MedFrameQA~\cite{yu2025medframeqa}, and BESTMVQA~\cite{hong2023bestmvqa} evaluate multimodal reasoning over radiology and pathology images.
However, these datasets typically assume clean OCR or use synthetic templates, and they do not address the structured interpretation of noisy, real-world medical report images.
Docopilot~\cite{duan2025docopilot} recently explored end-to-end document reasoning with layout-aware VLMs, but was not tailored to clinical interpretation or evaluation.
In contrast, MedRepBench emphasizes OCR robustness, layout variance, and medically grounded extraction, filling a key gap in VLM evaluation.

\subsection{LLMs for Structured Extraction and Evaluation}
LLMs have shown potential for extracting structured information from clinical texts~\cite{garcia2025langchainextract} and for evaluating generation tasks via automatic metrics~\cite{liu2023gptjudge}.
MedRepBench uses LLMs in both roles when OCR is available: as information extractors in OCR-assisted pipelines and as subjective evaluators for patient-facing explanations.
Unlike generic evaluation frameworks, our subjective evaluation emphasizes medical factuality and report-grounded interpretability, while remaining a diagnostic signal rather than an independent clinical assessment.
Moreover, recent analysis~\cite{zhang2025aesthetics} highlights the limitations of vision-only models in handling OCR-sensitive inputs, further justifying the dual no-OCR/OCR-assisted evaluation design in MedRepBench.

\section{Conclusion}

We present \textbf{MedRepBench}, a real-world benchmark for evaluating end-to-end, vision-grounded structured understanding of medical report images.
MedRepBench provides two complementary protocols: an \emph{objective} field-level recall metric for structured extraction and an \emph{automated subjective} framework that uses an LLM judge to assess report-grounded interpretability and factuality.
Benchmarking representative open-source models reveals that robust image-to-structure extraction remains challenging in the \textit{no-OCR} setting, leaving substantial headroom for future VLMs.
As a practical demonstration, we show that a lightweight RL-based alignment baseline using MedRepBench's objective reward can further improve a mid-scale VLM.

\section{Limitations} We mainly focus on open-source models to promote reproducibility and privacy-preserving evaluation. Therefore, we do not report results from proprietary models such as GPT, Gemini, and Claude.
MedRepBench is constructed from Chinese medical reports and does not claim cross-lingual or cross-country generalization, although the schema and evaluation protocols can be adapted to other languages and healthcare systems.
Our current baseline suite also emphasizes general-domain VLMs; medical-domain VLMs and additional OCR systems can be benchmarked through the released \href{https://huggingface.co/datasets/MedRepBench/MedRepBench}{resources}.

\bibliographystyle{splncs04}
\bibliography{main}
\end{document}